\documentclass[10pt,twocolumn,letterpaper]{article}

\usepackage[pagenumbers]{cvpr}      
\definecolor{cvprblue}{rgb}{0.21,0.49,0.74}
\usepackage[pagebackref,breaklinks,colorlinks,allcolors=cvprblue]{hyperref}

\def\confName{CVPR}
\def\confYear{2027}

\title{World Models Favor Learned Priors over Visual Evidence}

\author{
Han Xiao$^{1}$ \quad
Zhiyong Bao$^{2}$ \quad
Xiaotong Liu$^{3}$ \quad
Yanhao Chen$^{1}$ \quad
Qingqiang Wu$^{1}$
\\[8pt]
$^{1}$\,Xiamen University \quad
$^{2}$\,Northeastern University \quad
$^{3}$\,Tongji University
\\[6pt]
{\small
\texttt{cyhao@stu.xmu.edu.cn},
\texttt{wuqq@xmu.edu.cn}}
}

\begin{document}
\maketitle

\begin{abstract}
World models are increasingly used to simulate physical interactions, but training a separate model for each environment limits scalability. Cross-environment prediction requires identifying the current physical rule, which appearance alone may not reveal. We therefore study visual calibration: using object motion and interactions in a short video as direct evidence to infer the current rule and predict motion under that rule from a different initial state. However, differences in data availability across physical conditions can yield a long-tailed training distribution, biasing models toward common rules even when evidence supports rare ones. To study this challenge, we introduce TailWorld, which evaluates the transfer of demonstrated rules across initial states while independently controlling training imbalance and calibration informativeness. The benchmark compares predictions with two reference futures from the same prediction initial state, one following the demonstrated rule and the other a common rule, to measure prediction bias toward common dynamics. We further propose Rule Inference for Transfer (RIFT), which learns how different rules affect object motion, infers the current rule from calibration interactions, and applies it to prediction from a new initial state. RIFT explicitly controls the influence of training frequency on rule inference, allowing sufficiently informative observational evidence to support rare rules. Experiments on TailWorld show that RIFT improves prediction accuracy and physical consistency across varying training imbalances and distribution shifts, with gains in predicting rare dynamics.
\end{abstract}


\begin{figure}[t]
    \centering
    \includegraphics[width=\linewidth]{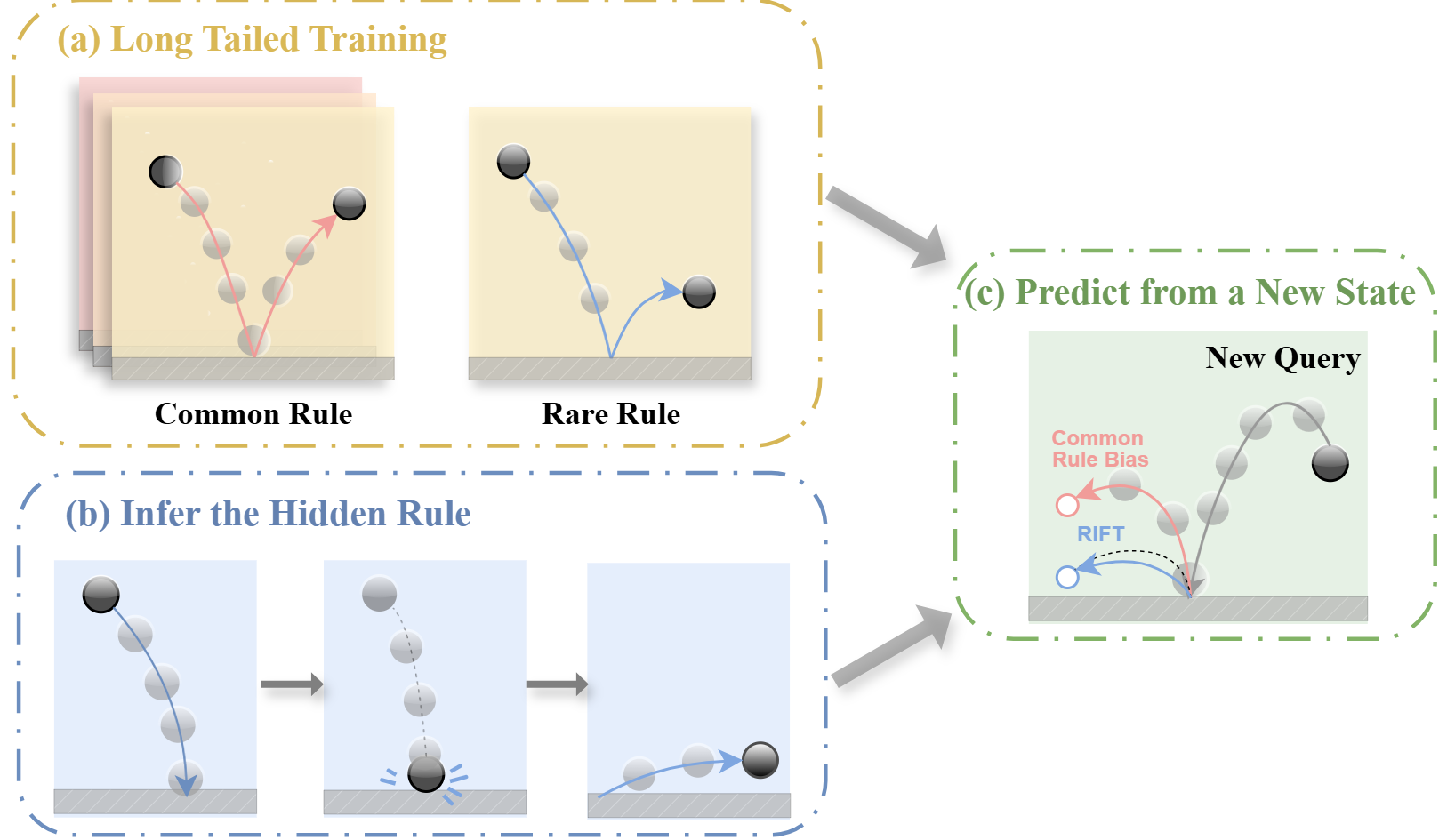}
    \caption{RIFT transfers calibrated physical rules across states to predict rare dynamics.}
    \label{fig:Intro}
\end{figure}

\section{Introduction}

World models are increasingly used to simulate how physical environments evolve \cite{ha2018recurrent,bruce2024genie,zhou2025dinowm}. Training a separate model for each environment is costly and difficult to scale, motivating world models that can predict across environments. A key challenge is identifying the physical rule governing the current environment. Visually similar objects may behave differently because their physical properties or surrounding conditions differ, so appearance alone may be insufficient \cite{wu2017deanimation,baradel2020cophy}. These properties and conditions shape motion and interactions \cite{battaglia2016interaction,sanchezgonzalez2020gns,pfaff2021mesh}, allowing the dynamics captured in a video to provide evidence for identifying the current rule \cite{riochet2018intphys,bear2021physion}. We study \emph{visual calibration}, where a model infers the current rule from a short demonstration and predicts motion under that rule from a different initial state \cite{baradel2020cophy}. Changing the initial state makes replaying the demonstrated trajectory insufficient, testing whether the model can understand and transfer the underlying rule.

Recent methods primarily adapt models to different dynamics by updating environment-specific parameters from limited observations \cite{kirchmeyer2022coda,koupai2024geps,nzoyem2025ncf} or conditioning predictions on representations of observed trajectories \cite{baradel2020cophy,dahlinger2025mango}. However, real-world objects, materials, and environments differ in prevalence and ease of data collection, leading to uneven coverage of physical conditions and potentially a long-tailed data distribution. Models trained on such data may predict according to common rules even when calibration evidence supports rare ones. The key challenge is whether a model can base its predictions on sufficiently informative observations when they conflict with preferences induced by training frequency.

To study this challenge under controlled conditions, we introduce \textbf{TailWorld}, a benchmark for visual calibration under long-tailed dynamics. Each task requires prediction from a different initial state under the rule demonstrated in a calibration video. TailWorld independently controls rule frequencies during training and calibration informativeness to examine how training imbalance and observational evidence jointly affect prediction. It also provides two reference futures from the same prediction initial state: one follows the demonstrated rule, and the other follows a common rule. Comparing predictions with both quantifies their deviation from the true future and any shift toward the common alternative, directly assessing the model's preference for common dynamics.

To enable models to infer rules from current evidence and apply them to new trajectories, we propose \textbf{Rule Inference for Transfer (RIFT)}. RIFT first learns how different physical rules affect object motion, then uses these relationships to infer the current rule from interactions observed in a calibration video. It applies the inferred rule to prediction from a different initial state and keeps it fixed throughout prediction. Crucially, RIFT explicitly controls the influence of training frequency on rule inference, allowing sufficiently informative observations to support rules rarely seen during training. The model can thus draw on knowledge learned across environments to predict motion under the current rule from a new initial state.
Experiments on TailWorld show that RIFT improves prediction accuracy and physical consistency across varying training imbalances and distribution shifts, with gains extending to rare dynamics. Our contributions are threefold:
\begin{enumerate}
    \item We study visual calibration under long-tailed dynamics, examining whether models can follow rare rules supported by observational evidence when predicting motion from different initial states.
    \item We introduce TailWorld, which independently controls training imbalance and calibration informativeness and measures prediction bias toward common dynamics using reference futures under the demonstrated and common rules.
    \item We propose RIFT, which infers the current rule from observed interactions, explicitly controls the influence of training frequency on inference, and applies the inferred rule to prediction from different initial states.
\end{enumerate}

\section{Related Work}
\label{sec:related_work}

\paragraph{World models and dynamics adaptation.}
World models predict environment evolution to support planning
and adaptation
\citep{ha2018recurrent,hafner2019planet,hafner2020dreamer,
micheli2023iris,alonso2024diamond,bruce2024genie,zhou2025dinowm}.
For physical systems, interaction-based models capture
object interactions and their effects on motion
\citep{battaglia2016interaction,chang2017compositional,
li2019particle,sanchezgonzalez2020gns,pfaff2021mesh}.
Adaptation methods update model parameters or condition
predictions on limited observations
\citep{finn2017maml,garnelo2018cnp,sanchezgonzalez2018graph,
kirchmeyer2022coda,koupai2024geps,nzoyem2025ncf},
while recent work extracts physical information from
trajectories and interaction videos
\citep{dahlinger2025mango,luo2026physicalworld}.
Under imbalanced training frequencies, however, predictions
may favor a common rule even when calibration evidence
supports a rare one.

\paragraph{Visual physical inference and evaluation.}
Visual physical inference recovers hidden properties and
interactions from observed motion
\citep{wu2015galileo,wu2016physics101,wu2017deanimation,
kipf2018nri,li2020causal}.
Counterfactual prediction tests whether inferred dynamics
transfer across initial conditions
\citep{baradel2020cophy,janny2022filteredcophy}.
Benchmarks evaluate physical plausibility, outcome prediction,
and causal reasoning
\citep{riochet2018intphys,piloto2022intuitive,bakhtin2019phyre,
yi2020clevrer,bear2021physion},
with further work probing hidden properties and
challenging interactions
\citep{tung2023physionpp,kang2025physical,bansal2025videophy,
motamed2026physicsiq,li2026tailor}.
Prediction error alone, however, does not distinguish
ambiguous evidence from a preference for common dynamics.
TailWorld independently varies training imbalance and
calibration informativeness, then compares predictions
from a new initial state with futures under the demonstrated
and common rules.
This reveals whether predictions favor common dynamics
despite evidence identifying a rare rule.

\paragraph{Long-tailed learning and prior correction.}
Long-tailed learning addresses target imbalance through
reweighting, margin adjustments, and decoupled training
\citep{cui2019classbalanced,cao2019ldam,kang2020decoupling}.
Prior-aware methods account for label frequencies and
distribution shifts
\citep{ren2020balanced,menon2021logit,hong2021lade,
lipton2018labelshift},
with extensions to continuous targets
\citep{yang2021imbalanced}.
These approaches motivate frequency correction, but physical
prediction requires inferring a latent rule from interactions
and transferring it to a new initial state.
RIFT addresses this coupled problem by learning how rules
shape motion, inferring the governing rule from calibration
video while controlling training-frequency effects, and
using that rule to predict future dynamics.

\section{TailWorld Benchmark}
\label{sec:benchmark}

To examine how calibration evidence and training frequency
influence prediction, TailWorld separates observations
that establish the query motion state from interactions
that reveal the governing rule.
Each task pairs a calibration video with a query trajectory
under the same rule but different initial positions and
velocities.
The model observes a collision-free query segment and
predicts subsequent motion, using calibration interactions
to distinguish rules that the query history alone cannot
identify.

\begin{figure}[t]
    \centering
    \includegraphics[width=\linewidth]{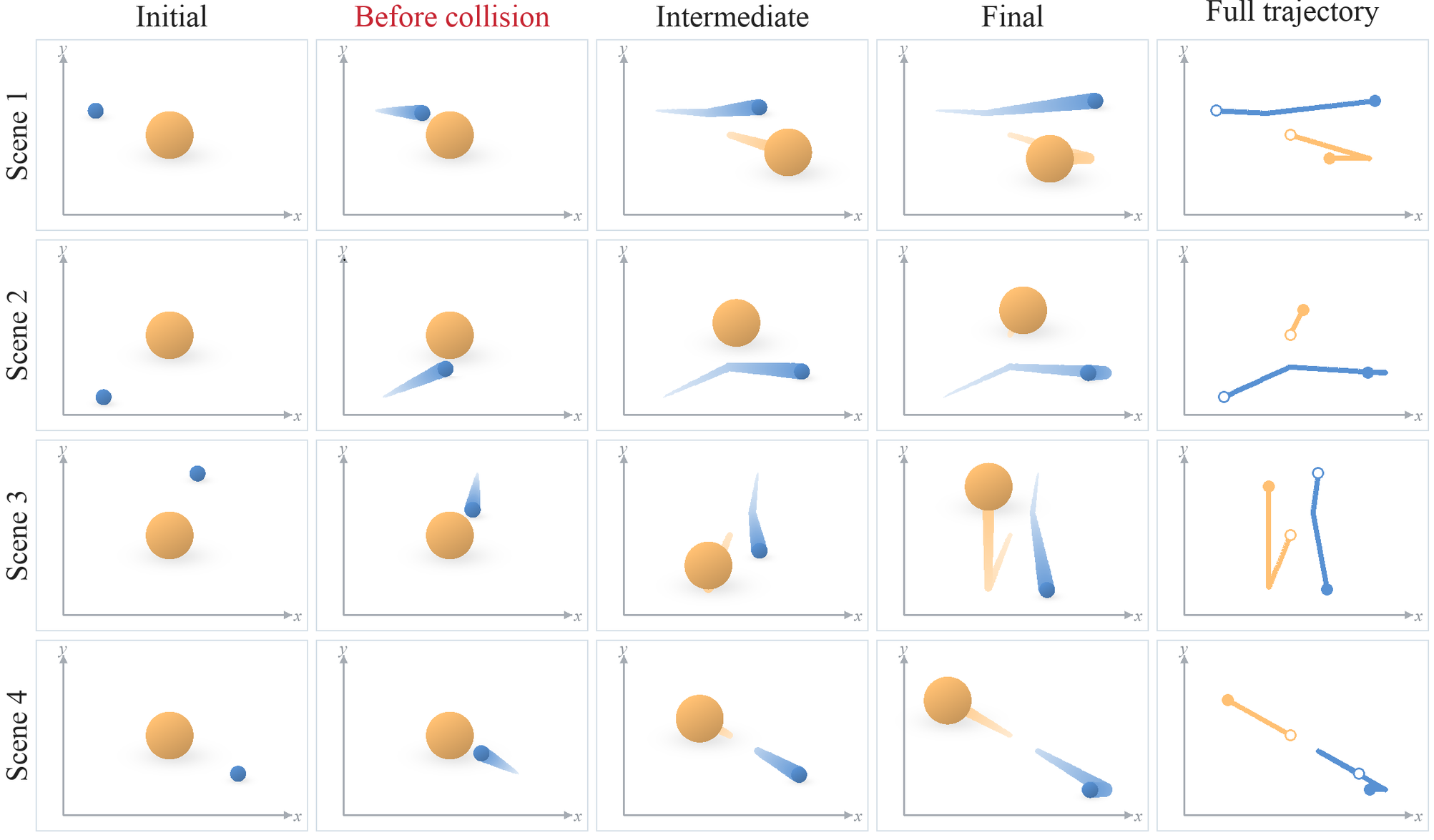}
    \caption{Examples of 2D scenes in TailWorld.}
    \label{fig:dataset1}
\end{figure}

\begin{figure}[t]
    \centering
    \includegraphics[width=\linewidth]{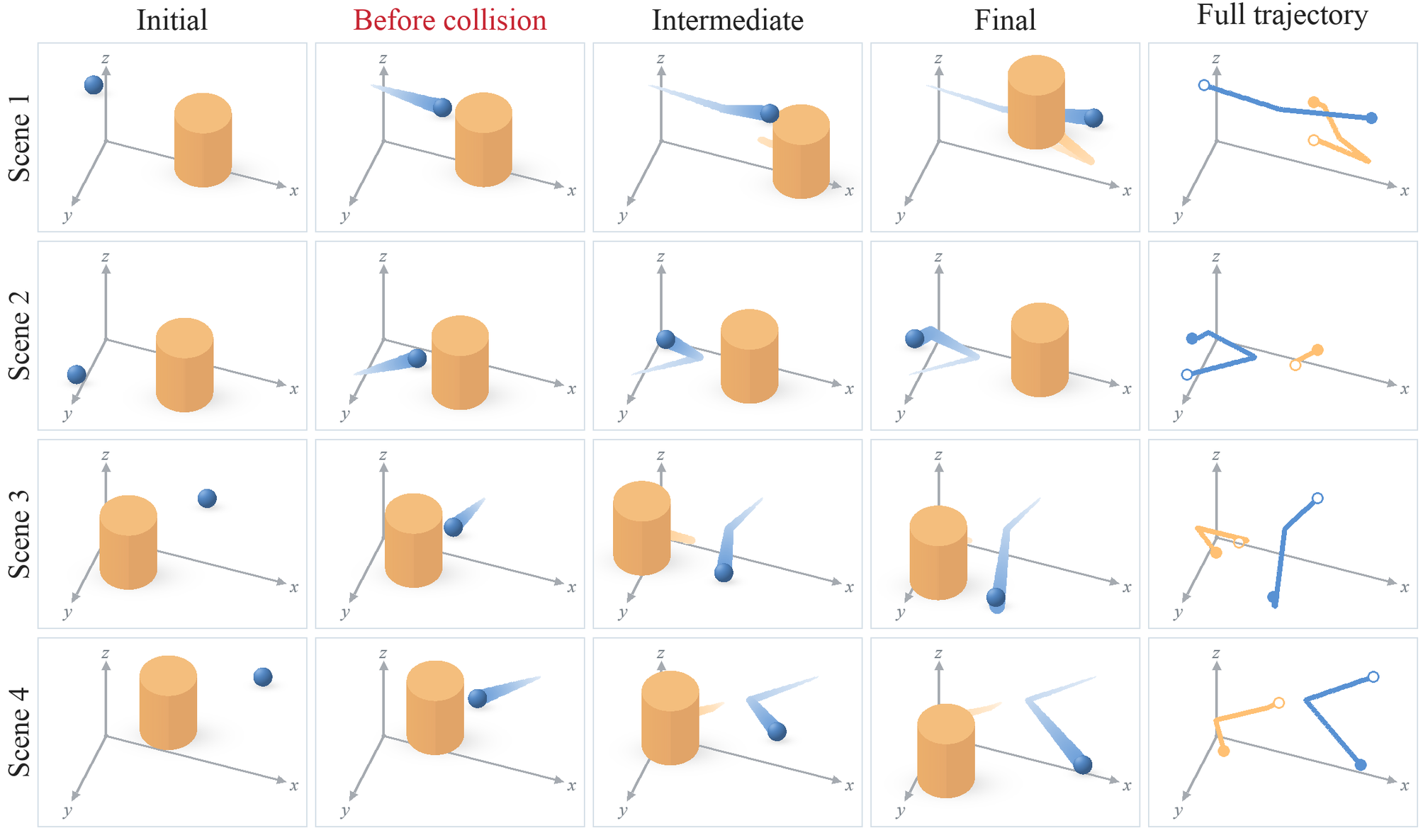}
    \caption{Examples of 3D scenes in TailWorld.}
    \label{fig:dataset2}
\end{figure}

We achieve this separation through collision scenes,
where restitution and mass ratio affect collision outcomes
but leave contact-free motion unchanged.
The 2D scenes contain a disc and a movable circular pillar
(Fig.~\ref{fig:dataset1});
the 3D scenes contain a sphere and a movable finite cylinder,
observed by three calibrated cameras
(Fig.~\ref{fig:dataset2}).
We vary restitution, mass ratio, or both while keeping
appearance and scene geometry fixed.
This construction makes calibration informative about
the governing rule and subsequent collisions informative
about which rule the prediction follows.

\subsection{Two Futures from the Same Initial State}

For each synthetic scene, we generate two future trajectories
from exactly the same initial state: one follows the rule
demonstrated in the calibration video, and the other follows
a reference rule common in training.
The model observes the first $0.5$\,s of query motion and
predicts both objects over the next 5\,s, corresponding to
15 observed and 150 predicted frames at 30\,Hz.
The observed segment must be collision-free and identical
under both rules.
It therefore reveals object positions and velocities
without distinguishing the rules, making calibration
evidence necessary for rule identification.
For samples assessing bias toward common rules, we further
require the two futures to differ clearly after subsequent
collisions, ensuring that the governing rule materially
affects prediction.
Both complete trajectories are withheld at inference
and serve as references for assessing whether predictions
more closely follow the calibration-supported or common rule.

\subsection{Varying Training Frequency and Available Evidence}

We adjust sampling proportions within a fixed training pool
to progressively increase common-rule frequency while
keeping rare-rule sampling probability unchanged.
Rare cases retain $5\%$ of the sampling probability,
with a common-to-rare sampling ratio of $r\!:\!1$,
where $r\in\{1,4,16\}$; other training cases receive
the remaining probability.
Under a fixed training budget, this preserves the expected
number of rare-rule exposures, allowing us to examine
the effect of increasing common-rule prevalence.
Separately, we provide shorter or longer segments of the
same calibration video to test whether additional motion
and interaction evidence aids rule identification.
The shorter segment ends $0.6$\,s after the first collision;
the longer contains the full 6\,s trajectory.
For each training distribution, we use both calibration
lengths on the same prediction tasks to examine the
respective effects of training distribution and current
video evidence on prediction.

\subsection{Testing Rule Transfer}

Beyond assessing prediction under rules infrequent in
training, we test whether rules transfer to initial states
not paired with them during training.
All test tasks combine rules and initial states that
have each appeared in training, but whose pairings
are excluded from training.

To assess transfer across initial states in real observations,
we use pendulum recordings from
IRIS~\cite{khanbayov2026iris} and
Delfys75~\cite{Garcia_2025_CVPR}.
We pair separate recordings under the same physical setting
as calibration and query videos, holding out selected
settings for testing.
These tasks use $0.5$\,s of query history and a 3\,s
prediction horizon.

\subsection{Demonstration Robustness and Control Cases}

We complement the transfer tests with two checks on how
calibration affects prediction.
First, we provide two distinct calibration trajectories
per rule, each used separately as input, to test whether
predictions remain consistent across demonstrations
of the same rule.
Second, we retain cases where the demonstrated and common
rules produce nearly identical futures, checking for
unnecessary prediction changes when rule differences
have little effect on the target motion.
Full parameters and construction details appear in Appendix~B.

\section{Methodology}
\label{sec:methodology}

\begin{figure*}[t]
    \centering
    \includegraphics[width=\textwidth]{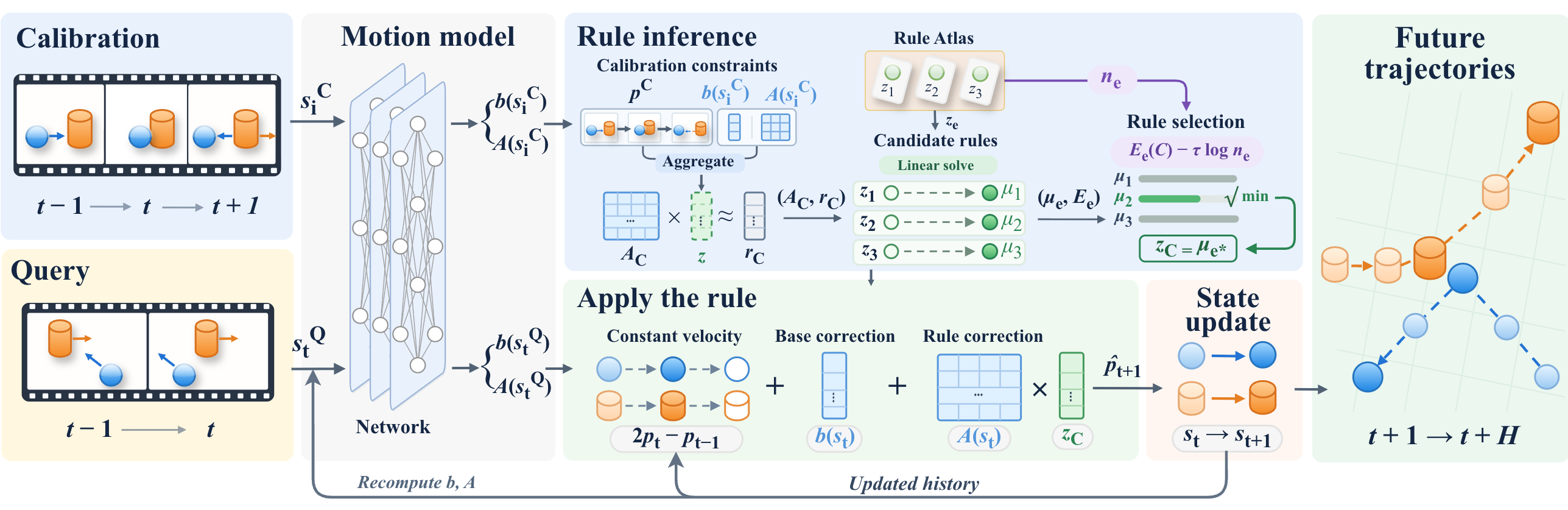}
    \caption{Overview of RIFT. RIFT infers physical rules from calibration interactions, explicitly controls the influence of training frequency on rule selection, and transfers the inferred rule to trajectory prediction from different initial states.}
    \label{fig:overview}
\end{figure*}

\subsection{Problem Formulation and Overview}

We study cross-state physical prediction under long-tailed training distributions. Each sample pairs a calibration demonstration $C$ with a query sequence $Q$ governed by the same physical rule but initialized independently. The goal is to infer this rule from observations in $C$ and predict the query's future trajectory from its initial observations $Q_{\mathrm{obs}}$.

Rule Inference for Transfer (RIFT) infers a rule from observed motion and applies it to prediction from a new initial state. A learned rule vector represents the environmental rule, shared across trajectories in the same environment. The motion model maps the current state to a base correction and a response matrix, making next-step position predictions linear in this vector. Calibration observations thus yield linear constraints for recovering it. During query prediction, the vector remains fixed while the state-dependent motion response evolves with the query motion.

To let calibration evidence support rare rules, RIFT selects candidates using separate explanatory cost and training frequency terms. Reference rule vectors fitted to training trajectories form a Rule Atlas. For each reference, RIFT balances calibration fit against deviation from that reference, yielding an updated candidate and its explanatory cost. Selection combines this cost with an independently adjustable frequency term. A rare candidate can be selected when its explanatory cost advantage outweighs the frequency preference for common candidates.

\subsection{Learning Motion Relationships for Rule Identification}

This module learns how rules affect motion while keeping predictions linear in the rule vector, enabling its recovery from calibration and reuse for prediction from different initial states. Let $p_t \in \mathbb{R}^{D}$ concatenate all object coordinates at time $t$, with total dimension $D$. For uniformly spaced frames, constant-velocity extrapolation gives $2p_t-p_{t-1}$. We learn a correction:
\begin{equation}
    \hat p_{t+1}
    =
    2p_t-p_{t-1}
    +
    b(s_t)+A(s_t)z.
    \tag{1}
\end{equation}
Here, $\hat p_{t+1}$ is the predicted position, $s_t$ contains local motion history and scene information, and $z \in \mathbb{R}^{K}$ is the rule vector. Given $s_t$, the network produces a base correction $b(s_t) \in \mathbb{R}^{D}$ and a response matrix $A(s_t) \in \mathbb{R}^{D \times K}$. The base correction is independent of $z$; each column of $A(s_t)$ describes the current-state motion response associated with one rule coefficient. Weighted by $z$, these responses form the rule-dependent correction $A(s_t)z$, which combines with $b(s_t)$ to correct the extrapolated position.

Equation~(1) allows a shared rule to produce different effects across states. For example, a collision rule induces motion changes that depend on relative positions and velocities. The state-dependent $A(s_t)$ captures this variation, while $z$ remains shared within the environment.

This linear form supports both prediction and rule identification with the same motion model. For prediction, the current state and $z$ determine the next position. For identification, calibration supplies positions at adjacent time steps, and the network computes $b(s_t)$ and $A(s_t)$ from the observed state. Subtracting the constant-velocity extrapolation and base correction from the observed next position yields a linear constraint on $z$. Because calibration shares one rule vector throughout, constraints across time steps can be combined to solve for $z$.

\subsection{Recovering the Current Rule from Calibration}

This module recovers the rule from observed motion. When calibration cannot determine all coefficients, rule vectors estimated from training environments provide supplementary reference information.

\paragraph{Turning interactions into constraints.}
Equation~(1) yields the calibration constraint
\[
    A(s_t^C)z
    \approx
    p_{t+1}^C-2p_t^C+p_{t-1}^C-b(s_t^C),
\]
where the superscript $C$ denotes calibration. The right-hand side subtracts the constant-velocity extrapolation and base correction from the observed next position; the left-hand side is the response a candidate rule should produce at that state. A single environmental rule governs calibration, so one $z$ should explain motion changes throughout the sequence. We therefore solve jointly across time steps, combining rule evidence from different states.

We aggregate constraints over neighboring time steps and weight them by residual reliability to reduce position-difference noise. The resulting system is $\mathcal A_C z \approx r_C$, where $\mathcal A_C$ collects the response matrices and $r_C$ the corresponding motion residuals. Thus, $\|\mathcal A_C z-r_C\|_2^2$ measures a candidate's calibration fitting error, with $\|\cdot\|_2$ denoting the Euclidean norm.

\paragraph{Using references to supplement limited observations.}
Calibration reveals rule information through the motions and interactions present in the sequence. If certain interactions are absent, observations alone may not determine the corresponding coefficients. To supplement this information, we fit a reference vector $z_e$ to independent trajectories from each training environment $e$, minimizing motion fitting error while constraining coefficient magnitude. These vectors form the Rule Atlas.

For each reference $z_e$, we balance calibration fit against reference deviation to obtain an updated candidate, with minimum objective value
\begin{equation}
    E_e(C)
    =
    \min_z
    \left[
        \frac{1}{2}\|\mathcal A_C z-r_C\|_2^2
        +
        \frac{1}{2}(z-z_e)^\top\Lambda_0(z-z_e)
    \right].
    \tag{2}
\end{equation}
Here, $E_e(C)$ is the explanatory cost associated with reference $e$, ${}^\top$ denotes transposition, and $\Lambda_0$ is a positive-definite constraint matrix shared across references. Its strength is determined by the variability of rule estimates from independent trajectories within the same environment. The first term fits observed motion; the second limits reference deviation. The fitting term drives updates along directions constrained by observations. Where changes in the rule vector have little effect on calibration fit, the reference constraint helps determine its values.

Let $\mu_e$ denote the minimizer in Eq.~(2). Setting the quadratic objective's gradient to zero yields
\begin{equation}
    \bigl(\mathcal A_C^\top\mathcal A_C+\Lambda_0\bigr)\mu_e
    =
    \mathcal A_C^\top r_C+\Lambda_0z_e.
    \tag{3}
\end{equation}
This system balances observational and reference constraints by their relative strengths: calibration drives well-constrained directions more strongly, while weakly constrained directions retain more reference information. Each Atlas entry thus yields an updated candidate that can deviate continuously from its reference.

\subsection{Controlling Frequency Preference and Transferring the Rule}

This module combines candidate explanatory costs with an adjustable training-frequency preference to select a rule, then holds it fixed for query prediction from a different initial state. After updating each reference, we select
\begin{equation}
    e^\star
    =
    \arg\min_e
    \left[E_e(C)-\tau\log n_e\right],
    \qquad
    z_C=\mu_{e^\star}.
    \tag{4}
\end{equation}
Here, $n_e>0$ counts training exposures to environment $e$, $\tau\geq0$ controls the frequency preference, $e^\star$ indexes the selected reference, and $z_C$ is the rule vector used for prediction. The explanatory cost $E_e(C)$ includes both the updated candidate's calibration fitting error and its reference deviation. The term $-\tau\log n_e$ favors more frequent references, corresponding to prior weights proportional to $n_e^\tau$.

Equation~(4) selects the reference with the lowest total score. A rare reference $\ell$ scores lower than a frequent reference $h$, with $n_\ell<n_h$, when
\[
    E_h(C)-E_\ell(C)
    >
    \tau\log\frac{n_h}{n_\ell}.
\]
Its explanatory cost advantage must therefore outweigh the frequency preference for the frequent reference. Distinguishing interactions alone need not favor the rare candidate; calibration supports it when observed motion yields a sufficient explanatory cost advantage. Reducing $\tau$ lowers this threshold. At $\tau=0$, all references have equal prior weights, and selection depends solely on Eq.~(2). Since frequency enters only through Eq.~(4), its influence can be adjusted without retraining the motion model or refitting the Atlas.

We initialize the motion history with $Q_{\mathrm{obs}}$ and recursively predict the query using Eq.~(1) with $z_C$. At each step, predicted positions update the state, and $b$ and $A$ are recomputed while $z_C$ remains fixed. The query's positions, velocities, and interactions thus evolve jointly under the calibration-inferred rule and the evolving query state.

\subsection{Training the Framework through Cross-State Prediction}

This module learns motion relationships through prediction from a different initial state, enabling a calibration-inferred rule to transfer to a new trajectory. During training, we sample calibration--query pairs from the same environment with independently sampled initial states. We first estimate the rule directly from calibration:
\begin{equation}
    \tilde z_C
    =
    \arg\min_z
    \left\{
        \|\mathcal A_C z-r_C\|_2^2
        +
        \lambda_C\|z\|_2^2
    \right\}.
    \tag{5}
\end{equation}
Here, $\tilde z_C$ is the training-time rule estimate, and $\lambda_C>0$ is the regularization strength. The observation term fits calibration, while regularization limits coefficient magnitudes along weakly constrained directions, stabilizing estimation under limited observations. The same procedure estimates Atlas references from independent trajectories in training environments.

We apply $\tilde z_C$ to the query, predict recursively from the end of its observation segment, and minimize future trajectory error:
\begin{equation}
    \mathcal L_{\mathrm{exec}}
    =
    \mathbb E_{e,C,Q}
    \left[
        \frac{1}{H}
        \sum_{k=1}^{H}
        \left\|
            \hat p_k^Q(\tilde z_C)-p_k^Q
        \right\|_2^2
    \right].
    \tag{6}
\end{equation}
The cross-state prediction loss $\mathcal L_{\mathrm{exec}}$ averages errors over a horizon of $H$ steps following the query observation segment. At step $k$, $\hat p_k^Q(\tilde z_C)$ denotes the prediction under the calibration-inferred rule, and $p_k^Q$ denotes the ground-truth position. The expectation is over training environments and independently initialized calibration--query pairs within each environment.

Training with Equation~(5) encourages calibration-based transfer without relying on Atlas references. Gradients through the solver and query prediction train $b$ and $A$ to make inferred coefficients transferable across states within the same environment. Inference retains the learned representation, observation constraints, and prediction mechanism, but replaces zero-centered regularization with regularization around Atlas references to guide estimation along directions weakly constrained by calibration observations. Theoretical analysis appears in Appendix~A, hyperparameter and efficiency analyses appear in Appendix~D.

\section{Experiment}

\subsection{Experimental Setup}

\paragraph{Data and protocols.}
We evaluate RIFT on the 2D and 3D collision scenes in TailWorld and assess its ability to predict real-world motion using pendulum videos from IRIS~\cite{khanbayov2026iris} and Delfys75~\cite{Garcia_2025_CVPR}. Each task requires predicting future motion from a different initial state under the same physical rule, given an independent calibration demonstration. The 2D and 3D models are trained and evaluated separately. In synthetic experiments, all trajectory methods share a frozen visual state readout $E_\eta$, trained separately for 2D and 3D, to extract states from both query and calibration videos within their permitted observation windows (Appendix~B.2). Real pendulum experiments use position trajectories extracted from video and calibrated geometrically.

We conduct two complementary comparisons. First, we compare RIFT with CoPhyNet~\cite{baradel2020cophy}, GEPS~\cite{koupai2024geps}, CoDA~\cite{kirchmeyer2022coda}, NCF~\cite{nzoyem2025ncf}, and MaNGO~\cite{dahlinger2025mango} to assess how effectively each method uses demonstrations to adapt to the current dynamics and transfer to new initial states. All methods use the same task splits, state observations, and demonstration durations. To compare adaptation mechanisms, RIFT and the shared-backbone variants of GEPS, CoDA, and NCF use the same pretrained, frozen object-interaction Transformer, with their respective prediction and adaptation modules trained separately. CoPhyNet and MaNGO retain their own dynamics architectures to compare the predictive performance of the complete methods.

Second, we include strong pretrained world models, Cosmos-Predict2.5-2B~\cite{nvidia2025worldsimulationvideofoundation} and Wan2.2-TI2V-5B~\cite{wan2025}, to assess their physical prediction capabilities on these tasks. Both models are evaluated without fine-tuning: on synthetic tasks, we compare their native inputs without demonstrations against inputs augmented with visual demonstrations; on real-video tasks, we use their native inputs. We extract trajectories from generated videos and score them using the same metrics, reporting trajectory recovery rates and the corresponding evaluation subsets.

\begin{table*}[t]
    \centering
    \caption{Prediction accuracy and physical consistency on TailWorld.}
    \label{tab:main_results}
    \renewcommand{\arraystretch}{0.90}
    \setlength{\tabcolsep}{6pt}
    \resizebox{\textwidth}{!}{%
    \begin{tabular}{lccccc}
        \toprule
        \textbf{Method}
        & \textbf{ADE} $\downarrow$
        & \textbf{FDE} $\downarrow$
        & \textbf{Velocity RMSE} $\downarrow$
        & \textbf{Kinetic Energy MAE} $\downarrow$
        & \textbf{Momentum MAE} $\downarrow$ \\
        \midrule
        \multicolumn{6}{c}{\textit{2D Physical Rule Transfer}} \\
        \midrule

        CoPhyNet~\cite{baradel2020cophy}
        & 0.1900 & 0.3980 & 0.1126 & 0.0146
        & 0.1168 \\

        GEPS~\cite{koupai2024geps}
        & 0.2201 & 0.4992 & \underline{0.0933} & 0.0200
        & \textbf{0.0879} \\

        CoDA~\cite{kirchmeyer2022coda}
        & 0.2612 & 0.5235 & 0.1571 & 0.0271
        & 0.1550 \\

        NCF~\cite{nzoyem2025ncf}
        & 0.2993 & 0.5554 & 0.1195 & 0.0376
        & 0.1376 \\

        MaNGO~\cite{dahlinger2025mango}
        & \underline{0.1767} & \underline{0.3485}
        & 0.1655 & \underline{0.0123} & 0.1696 \\

        Cosmos-Predict2.5-2B~\cite{nvidia2025worldsimulationvideofoundation}
        & 0.4375 & 0.6398 & 0.2356 & 0.0673
        & 0.4850 \\

        Wan2.2-TI2V-5B~\cite{wan2025}
        & 0.4693 & 0.6331 & 0.2631 & 0.1044
        & 0.4735 \\

        \textbf{RIFT}
        & \textbf{0.0895} & \textbf{0.1388}
        & \textbf{0.0722} & \textbf{0.0115}
        & \underline{0.1033} \\

        \midrule
        \multicolumn{6}{c}{\textit{3D Physical Rule Transfer}} \\
        \midrule

        CoPhyNet~\cite{baradel2020cophy}
        & 0.2023 & 0.4167 & 0.1483 & 0.0137
        & 0.1390 \\

        GEPS~\cite{koupai2024geps}
        & 0.1916 & \underline{0.3261}
        & \underline{0.1148} & \textbf{0.0057}
        & 0.1242 \\

        CoDA~\cite{kirchmeyer2022coda}
        & 0.1927 & 0.3908 & 0.1409 & 0.0234
        & \underline{0.0848} \\

        NCF~\cite{nzoyem2025ncf}
        & \underline{0.1772} & 0.3918
        & 0.1335 & 0.0188 & 0.1635 \\

        MaNGO~\cite{dahlinger2025mango}
        & 0.2081 & 0.3767 & 0.1667 & 0.0270
        & 0.1491 \\

        Cosmos-Predict2.5-2B~\cite{nvidia2025worldsimulationvideofoundation}
        & 0.5874 & 0.7565 & 0.3622 & 0.1332
        & 0.6641 \\

        Wan2.2-TI2V-5B~\cite{wan2025}
        & 0.6859 & 1.1701 & 0.3925 & 0.1058
        & 0.6513 \\

        \textbf{RIFT}
        & \textbf{0.0989} & \textbf{0.1205}
        & \textbf{0.0982} & \underline{0.0059}
        & \textbf{0.0689} \\

        \bottomrule
    \end{tabular}%
    }
\end{table*}

\paragraph{Metrics.}
We use five metrics to quantify prediction errors in position, velocity, and physical quantities. Average Displacement Error (ADE) averages the Euclidean distance between predicted and ground-truth positions over all objects and forecast steps. Final Displacement Error (FDE) measures the average position error across objects at the final forecast step. Both ADE and FDE are measured in meters. Velocity RMSE is the square root of the mean squared Euclidean norm of the velocity error over all objects and forecast steps. Predicted and ground-truth velocities are computed from their respective position trajectories using the same finite-difference scheme.

For synthetic scenes, we additionally report Kinetic Energy MAE and Momentum MAE. The total kinetic energy and the total linear momentum used in this work are defined as
\begin{equation}
K_t = \frac{1}{2}\sum_o m_o\|v_{to}\|_2^2,
\qquad
\mathbf{P}_t = \sum_o m_o v_{to},
\end{equation}
where $m_o$ and $v_{to}$ denote object mass and velocity, respectively. Both metrics sum quantities over objects, then compute errors between predicted and ground-truth totals using absolute differences for energy and Euclidean distances for momentum, averaging across forecast steps. Lower values indicate better performance for all five metrics.

\begin{figure}[t]
    \centering
    \includegraphics[width=\linewidth]{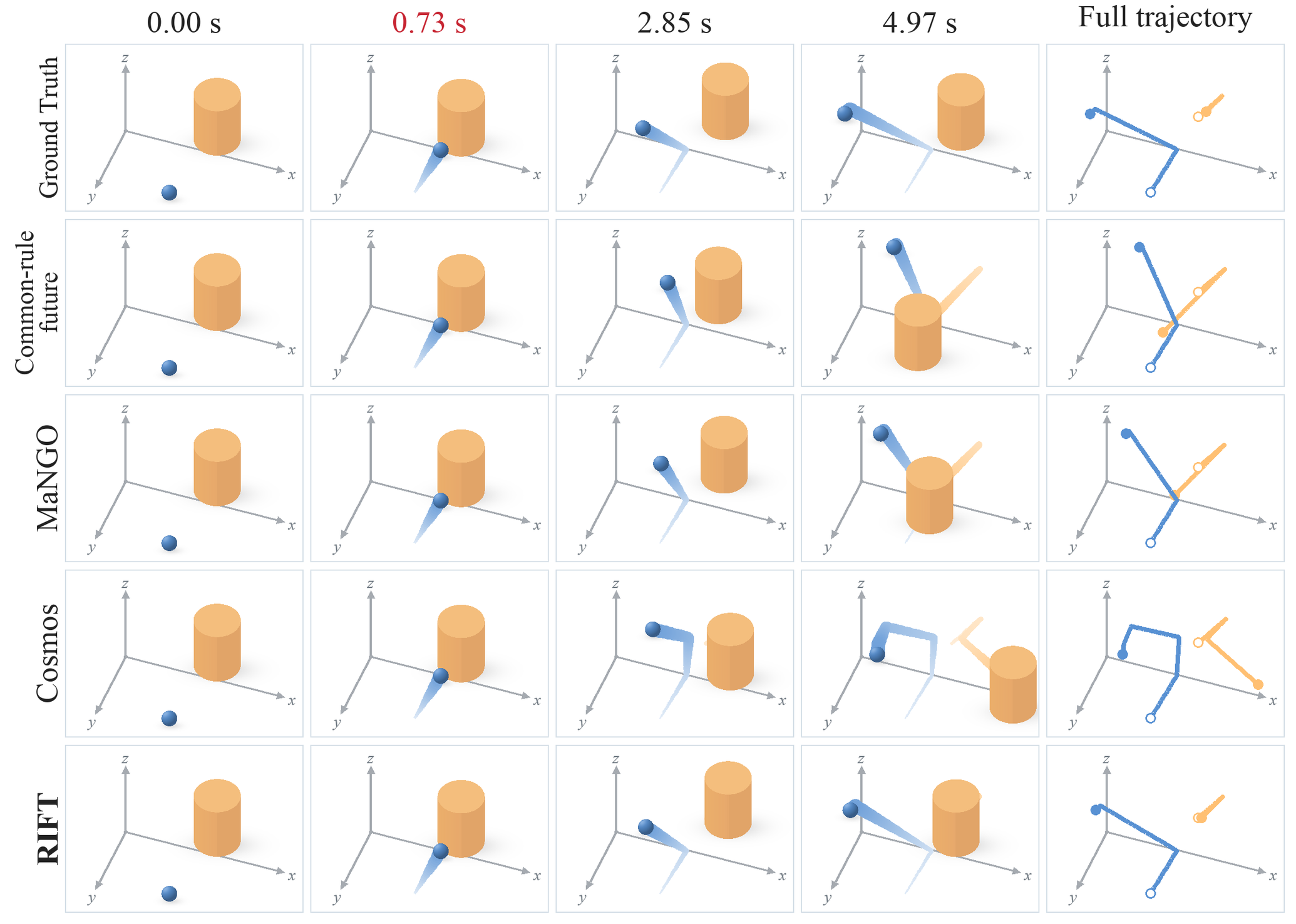}
    \caption{Trajectory prediction under long-tailed dynamics, where the ground truth trajectories exhibit rare dynamics..}
    \label{fig:tail}
\end{figure}

\subsection{Trajectory Accuracy and Physical Consistency}
As shown in Table~\ref{tab:main_results}, RIFT achieves the best trajectory and velocity predictions in both 2D and 3D physical rule transfer tasks, demonstrating consistent advantages across settings. Joint reductions in trajectory and velocity errors show that these advantages extend from object positions to motion evolution. RIFT also leads in 2D kinetic energy and 3D momentum prediction, providing complementary evidence of physical consistency. Together, these results demonstrate RIFT's ability to combine accurate trajectory prediction with faithful physical quantity estimates under physical rule transfer. Appendix~C.1 shows that RIFT maintains competitive accuracy across different demonstrations of the same rule.

\subsection{Physical Rule Transfer under Training Imbalance}
Figure~\ref{fig:tail} compares predicted and ground-truth trajectories, with the common-rule trajectories from the same initial state serving as a reference. We select MaNGO as a competitive baseline and also include Cosmos. MaNGO's post-collision predictions more closely follow the common-rule reference, indicating a bias toward the frequent rule, while Cosmos also fails to accurately reproduce the true collision response. RIFT closely follows the ground truth throughout the rollout, accurately capturing the ball's deflection and the cylinder's limited displacement. These results show that RIFT uses demonstrations to predict motion consistent with the current rule, even when it differs from the pattern prevalent in training. Appendix~C.2 shows that full calibration reduces RIFT's sensitivity to explicit preference for frequent rules.

\subsection{Rule Switching Guided by Demonstrations}
\begin{figure}[t]
    \centering
    \includegraphics[width=\linewidth]{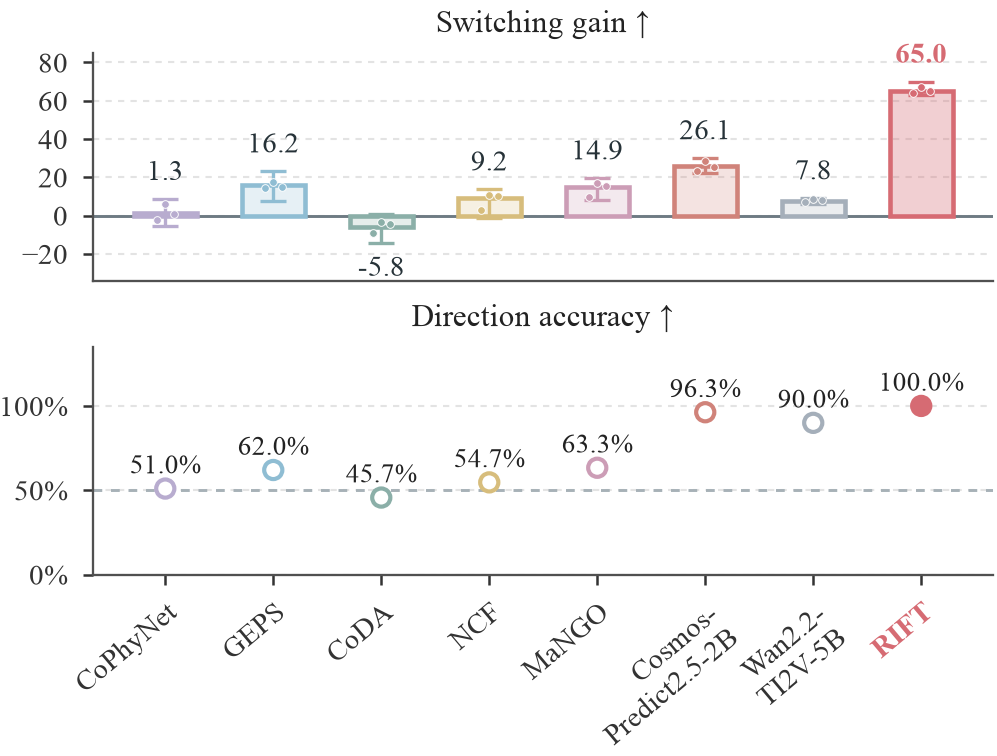}
    \caption{Rule switching guided by calibration demonstrations.}
    \label{fig:switch}
\end{figure}

We fix the query initial state and vary only the physical rule in the calibration demonstration to test whether predictions change accordingly. In Figure~\ref{fig:switch}, switching gain measures the reduction in trajectory error achieved by correct rather than incorrect demonstrations; direction accuracy measures the fraction of samples whose predicted trajectories shift in the correct direction after changing the demonstration. RIFT leads on both metrics, achieving the largest error reduction and correct directional shifts across all valid samples. Baselines do not attain comparable performance on both metrics simultaneously. By combining correct directional shifts with substantial accuracy gains, RIFT demonstrates its ability to use rule information from demonstrations to adjust future predictions. Appendix~C.3 confirms that RIFT also leads in switching gain and direction accuracy in 3D.

\subsection{Effect of Calibration Evidence on Rule Transfer}
\begin{figure}[t]
    \centering
    \includegraphics[width=\linewidth]{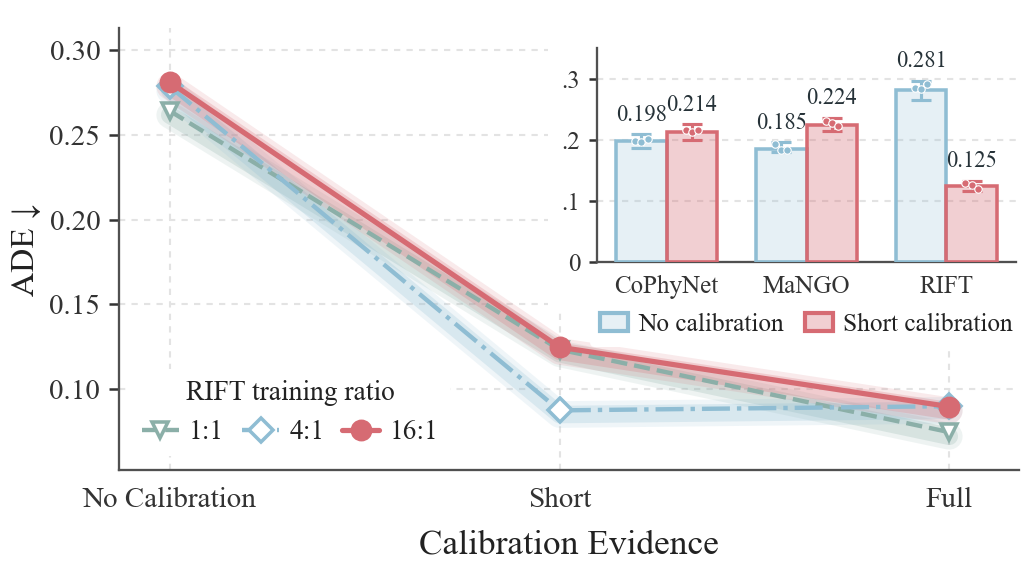}
    \caption{Effect of calibration evidence on trajectory prediction under varying training imbalance.}
    \label{fig:evidence}
\end{figure}

We evaluate RIFT with no, short, and full demonstrations under different training frequency ratios to examine how it uses demonstrations to improve prediction. As shown in Figure~\ref{fig:evidence}, short demonstrations substantially reduce trajectory errors across all training settings, while full demonstrations yield low errors across training ratios. The inset further shows that, although RIFT has higher errors without demonstrations, the gains from short demonstrations enable it to outperform the baselines shown. These results demonstrate that RIFT effectively uses limited calibration evidence to achieve accurate predictions under imbalanced training distributions. Appendix~C.4 confirms RIFT's lowest 3D ADE with short and full demonstrations across evaluated training ratios.

\begin{figure}[t]
    \centering
    \includegraphics[
        width=0.8\linewidth,
    ]{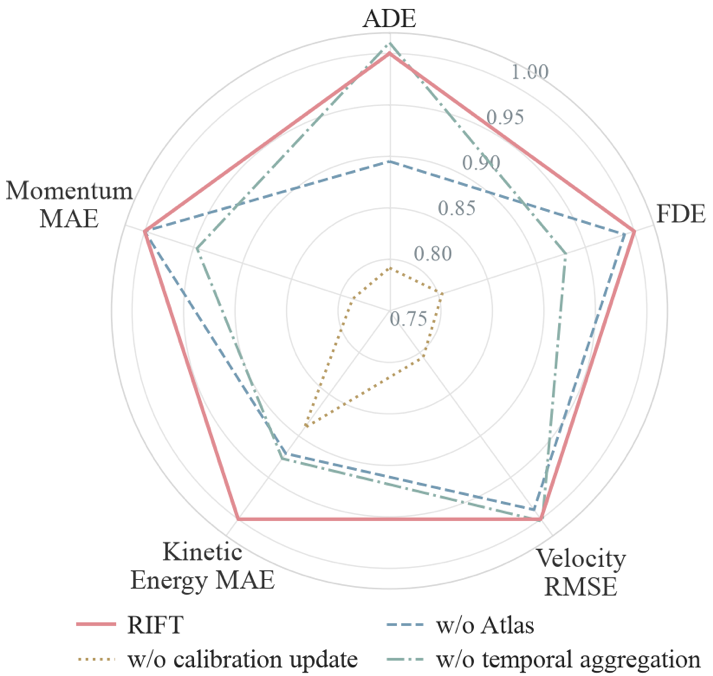}
    \caption{Ablation study of the key components in RIFT.}
    \label{fig:ablation}
\end{figure}

\begin{figure}[t]
    \centering
    \includegraphics[width=\linewidth]{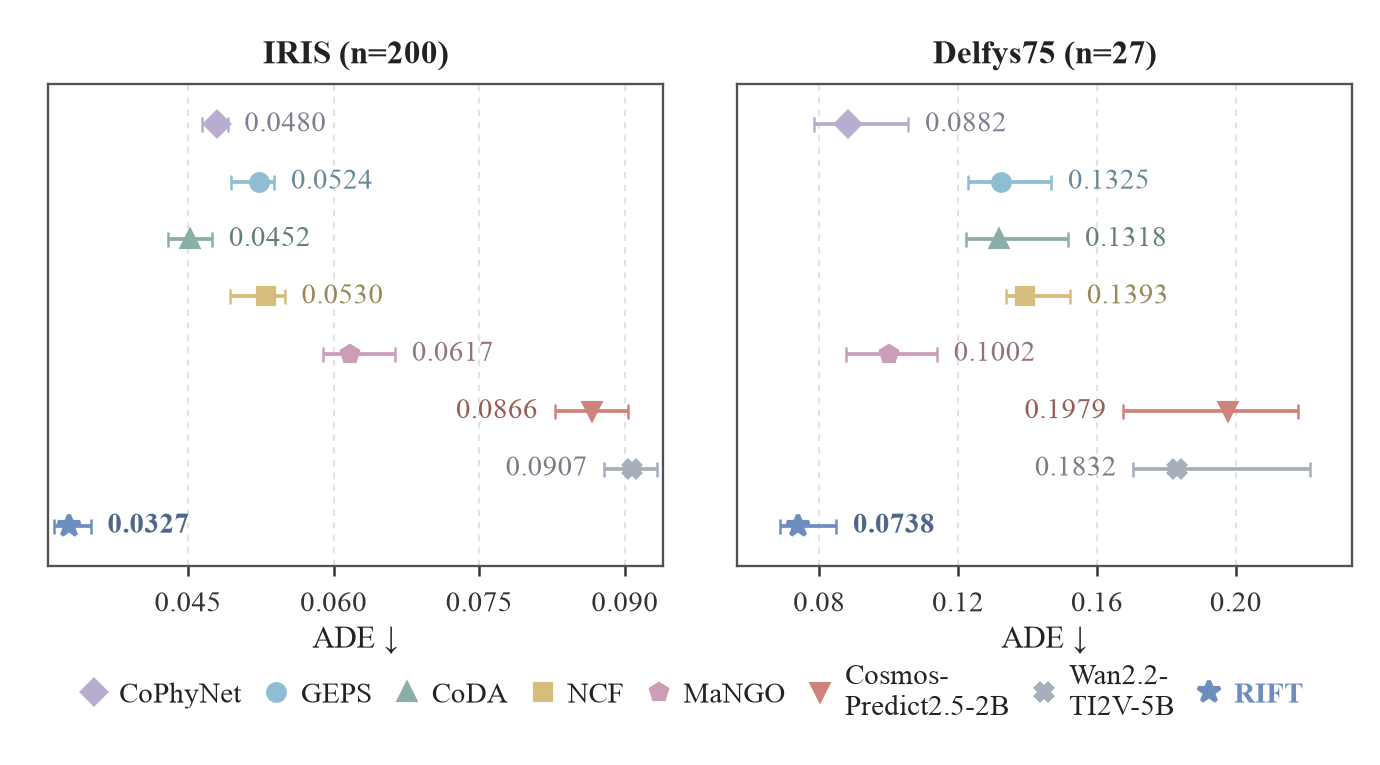}
    \caption{Trajectory prediction on real videos.}
    \label{fig:cross}
\end{figure}

\subsection{Trajectory Prediction in Real Videos}
We evaluate RIFT on the IRIS and Delfys75 real video datasets to assess its predictive capability under real observation conditions. As shown in Figure~\ref{fig:cross}, RIFT achieves the lowest ADE on both datasets, outperforming all baselines. This consistent advantage demonstrates that RIFT's trajectory prediction capability extends beyond synthetic scenes to real videos, providing further evidence of its effectiveness in real settings. Appendix~C.5 shows that RIFT also achieves the lowest ADE under parameter shift and unseen rule--state combinations.

\subsection{Ablation of Key Components}
Figure~\ref{fig:ablation} isolates the contributions of calibration updates, the Rule Atlas, and temporal aggregation. Removing calibration updates causes the largest performance drop, highlighting the importance of demonstration-driven rule refinement. The Rule Atlas supports prediction accuracy and physical consistency, while temporal aggregation improves selected physical metrics. Together, these components yield balanced performance across metrics. Appendix~C.6 further supports the value of rule updates and Atlas references under short calibration.

\section{Conclusion}
Predicting rare physical dynamics requires reliable rule inference that responds to calibration evidence despite imbalanced training. We introduced TailWorld to independently control training imbalance and calibration informativeness, enabling systematic evaluation of common-rule bias. RIFT infers governing rules from calibration observations and applies them to predict physical dynamics from new initial states. Its key advantage is explicit control over training frequency's influence on rule selection, enabling calibration evidence to guide the identification of rules rarely observed during training. Experiments demonstrate improved rare-dynamics prediction and lower errors on real videos. These results demonstrate that grounding rule inference in calibration evidence while controlling training-frequency influence can effectively mitigate common-rule bias and improve the prediction of rare physical dynamics.

\section*{LLM Usage Statement}
OpenAI's ChatGPT/GPT was used at an assistive level to review and refine author-produced text, related work and citations, code, experimental design and analysis, figures, data preparation, and research ideas, hypotheses, claims, and conclusions. It also provided reviewer-style feedback on the manuscript. The authors take full responsibility for all submitted content.

\bibliographystyle{ieeenat_fullname}
\bibliography{main}

\end{document}